\pdfoutput=1

\documentclass[11pt]{article}

\usepackage[]{acl}

\usepackage{times}
\usepackage{latexsym}

\usepackage[T1]{fontenc}

\usepackage[utf8]{inputenc}

\usepackage{microtype}
\usepackage{graphicx}
\usepackage{natbib}
\usepackage{doi}
\usepackage{paracol}
\usepackage{multirow}
\usepackage{xcolor}
\usepackage{amsmath}
\usepackage{booktabs}
\usepackage{amsfonts}
\usepackage{amssymb}
\usepackage{enumitem}
\title{Multilingual Molecular Representation Learning via Contrastive Pre-training}

\author{Zhihui Guo, Pramod Sharma, Andy Martinez, Liang Du, Robin Abraham \\
         Microsoft Corporation \\
         \small \texttt{\{zhihui.guo,pramod.sharma,andy.martinez,liang.du,robin.abraham\}@microsoft.com}}

\begin{document}
\maketitle
\begin{abstract}
Molecular representation learning plays an essential role in cheminformatics. Recently, language model-based approaches have gained popularity as an alternative to traditional expert-designed features to encode molecules. However, these approaches only utilize a single molecular language for representation learning. Motivated by the fact that a given molecule can be described using different languages such as Simplified Molecular Line Entry System (SMILES), the International Union of Pure and Applied Chemistry (IUPAC), and the IUPAC International Chemical Identifier (InChI), we propose a multilingual molecular embedding generation approach called MM-Deacon  (\textbf{m}ultilingual \textbf{m}olecular \textbf{d}omain \textbf{e}mbedding \textbf{a}nalysis via \textbf{con}trastive learning). MM-Deacon is pre-trained using SMILES and IUPAC as two different languages on large-scale molecules. 
We evaluated the robustness of our method on seven molecular property prediction tasks from MoleculeNet benchmark, zero-shot cross-lingual retrieval, and a drug-drug interaction prediction task.
\end{abstract}

\section{Introduction}
Drug discovery process involves screening of millions of compounds in the early stages of drug design, which is time consuming and expensive. Computer-aided drug discovery can  reduce the time and cost involved in this process via automating various cheminformatics tasks \cite{kontogeorgis2004computer,xu2017seq2seq,winter2019learning}.  

Traditional methods to encode molecules such as fingerprint generation rely heavily on molecular fragment-level operations on top of molecule graph constructed by molecular atoms and bonds \cite{burden1989molecular,bender2004molecular,vogt2008bayesian,muegge2016overview}. An example of such methods is Morgan fingerprint, also known as Extended-Connectivity Fingerprint (ECFP) \cite{morgan1965generation,rogers2010extended}, where a fixed binary hash function is applied on each atom and its neighborhood. These kinds of approaches focus on local features, hence they may not capture global information. 

\begin{figure}[t]
\centering
\includegraphics[width=0.48\textwidth]{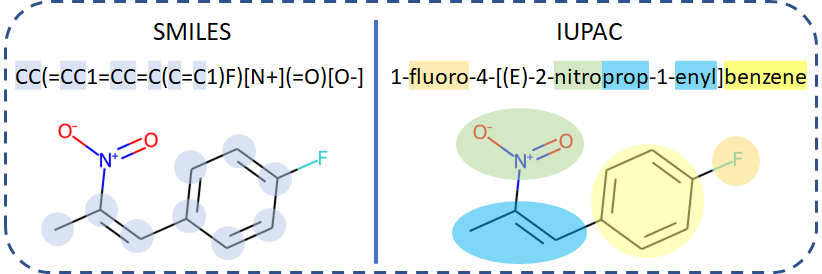}
\caption{Schematic illustration of differences in SMILES and IUPAC characteristics for the same molecule. 2D molecule graphs are plotted using RDKit \cite{landrum2013rdkit}. For the sake of simplicity, we only highlight atom \textit{C} in both SMILES string and the molecule graph below it. For IUPAC, we use the same color to denote the correspondence between the textual description in the string and the substructure on the molecule graph, where orange represents \textit{fluoro-} substituent, yellow represents \textit{benz-ene}, light green represents \textit{nitro-}, and turquoise represents \textit{prop-en-yl}. \label{fig:spdiff}}
\end{figure}

In addition to molecule graph, a given molecule can also be described with different languages such as Simplified Molecular Line Entry System (SMILES), the International Union of Pure and Applied Chemistry (IUPAC), and the IUPAC International Chemical Identifier (InChI). Particularly, SMILES is widely used to represent molecule structures as ASCII strings \cite{weininger1988smiles,favre2013nomenclature} at an atom and bond level. IUPAC nomenclature, on the other hand, serves the purpose of systematically naming organic compounds by basic words that indicate the structure of the compound and prioritize on functional groups to facilitate communication \cite{panico1993guide}. Fig. \ref{fig:spdiff} shows a comparison of SMILES and IUPAC characteristics for the same molecule. The SMILES string is created by traversing the molecule graph, where each letter in the SMILES string (such as \textit{C, F, N, O} in Fig. \ref{fig:spdiff}) corresponds to an atom on the graph, and other characters represent positions and connectivity. However, IUPAC names are akin to a natural language, and morphemes in the IUPAC name (like \textit{fluoro, prop, en, yl} in this example) often represent specific types of substructure on the molecule graph, which are also responsible for characteristic chemical reactions of molecules.  

Advances in natural language processing (NLP) have been very promising for molecule embedding generation and molecular property prediction \cite{xu2017seq2seq, gomez2018automatic,samanta2020vae,koge2021embedding,honda2019smiles,shrivastava2021fragnet, goh2017smiles2vec,schwaller2019molecular,payne2020bert,aumentado2018latent}. It is important to note that all of the methods mentioned above work with SMILES representation only. Therefore, the underlying chemical knowledge encoded in the embedding is restricted to a single language modality. Transformer models trained with self-supervised masked language modeling (MLM) loss \cite{vaswani2017attention} in chemical domain \cite{wang2019smiles,chithrananda2020chemberta,elnaggar2020prottrans,rong2020self,schwaller2021mapping,bagal2021liggpt} have also been used for molecular representation learning. However, pre-training objectives like MLM loss tend to impose task-specific bias on the final layers of Transformers \cite{carlsson2020semantic}, limiting the generalization of the embeddings.

In recent years, contrastive learning has been successful in multimodal vision and language research \cite{radford2021learning,meyer2020improving,shi2020contrastive,cui2020unsupervised,chen2021multimodal,alayrac2020self,akbari2021vatt,lee2020parameter,liu2020p4contrast}. Radford \textit{et al.} \citeyearpar{radford2021learning} used image-text pairs to learn scalable visual representations. Carlsson \textit{et al.} \citeyearpar{carlsson2020semantic} showed the superiority of contrastive objectives in acquiring global (not fragment-level) semantic representations. 

In light of these advances, we propose MM-Deacon (\textbf{m}ultilingual \textbf{m}olecular \textbf{d}omain \textbf{e}mbedding \textbf{a}nalysis via \textbf{con}trastive learning), a molecular representation learning algorithm built on SMILES and IUPAC joint training. Transformers are used as base encoders in MM-Deacon to encode SMILES and IUPAC, and embeddings from encoders are projected to a joint embedding space. Afterwards, a contrastive objective is used to push  the embeddings of positive cross-lingual pairs (SMILES and IUPAC for the same molecule) closer together and the embeddings of negative cross-lingual pairs (SMILES and IUPAC for different molecules) farther apart. Here instead of using SMILES and IUPAC for sequence-to-sequence translation \cite{rajan2021stout,krasnov2021struct2iupac,handsel2021translating}, we obtain positive and negative SMILES-IUPAC pairs and contrast their embeddings at the global molecule level rather than the fragment level. Different molecule descriptors are thus integrated into the same joint embedding space, with mutual information maximized across distinct molecule languages.

We pre-train MM-Deacon on 10 million molecules chosen at random from the publicly available PubChem dataset \cite{kim2016pubchem} and then use the pre-trained model for downstream tasks. Our main contributions are as follows:
\begin{itemize}
    \item We propose MM-Deacon, a novel approach for utilizing multiple molecular languages to generate molecule embeddings via contrastive learning.
    \item To the best of our knowledge, we are the first to leverage mutual information shared across SMILES and IUPAC for molecule encoding. 
    \item We conduct extensive experiments on a variety of tasks, including molecular property prediction, cross-lingual molecule retrieval, and drug-drug interaction (DDI) prediction, and demonstrate that our approach outperforms baseline methods and existing state-of-the-art approaches.
    
\end{itemize}

\begin{section}{Molecule pre-training}
Deep learning tasks commonly face two challenges: first, dataset size is often limited, and second, annotations are scarce and expensive. A pre-training scheme can benefit downstream tasks by leveraging large-scale unlabeled or weakly labeled data. Such pre-training and fine-tuning frameworks have recently sparked much interest in the molecular domain \cite{hu2019strategies, samanta2020vae,chithrananda2020chemberta, rong2020self, shrivastava2021fragnet, xue2021x, zhu2021dual, wang2021molclr, liu2021pre}. Existing pre-training methods can be divided into three categories based on the models used: pre-training with graph neural networks (GNNs), pre-training with language models, and pre-training with hybrid models.

\vspace{1mm}
\noindent \textbf{Pre-training with GNNs.} GNNs are a popular choice for molecule encoding that regard atoms as nodes and bonds as edges. Hu \textit{et al.} \citeyearpar{hu2019strategies} pre-trained GNNs on 2 million molecules using both node-level and graph-level representations with attribute masking and structure prediction objectives. MolCLR \cite{wang2021molclr} used subgraph-level molecule data augmentation scheme to create positive and negative pairs and contrastive learning to distinguish positive from negative. GraphMVP \cite{liu2021pre} was pre-trained on the consistency of 2D and 3D molecule graphs (3D graphs formed by adding atom spatial positions to 2D graphs) and contrastive objectives with GNNs.

\vspace{1mm}
\noindent \textbf{Pre-training with language models.} Language models are widely used to encode SMILES for molecular representation learning. Xu \textit{et al.} \citeyearpar{xu2017seq2seq} reconstructed SMILES using encoder-decoder gated recurrent units (GRUs) with seq2seq loss, where embeddings in the latent space were used for downstream molecular property prediction. Chemberta \cite{chithrananda2020chemberta} fed SMILES into Transformers, which were then optimized by MLM loss. FragNet \cite{shrivastava2021fragnet} used encoder-decoder Transformers to reconstruct SMILES and enforced extra supervision to the latent space with augmented SMILES and contrastive learning. X-Mol \cite{xue2021x} was pretrained by taking as input a pair of SMILES variants for the same molecule and generating one of the two input SMILES as output with Transformers on 1.1 billion molecules.

\vspace{1mm}
\noindent \textbf{Pre-training with hybrid models.} Different molecule data formats can be used collaboratively to enforce cross-modality alignment, resulting in the use of hybrid models. For example, DMP \cite{zhu2021dual} was built on the consistency of SMILES and 2D molecule graphs, with SMILES encoded by Transformers and 2D molecule graphs encoded by GNNs.
\end{section} 

Unlike other molecule pre-training methods, MM-Deacon is multilingually pre-trained with language models using pairwise SMILES and IUPAC. Compared with using molecule graphs with GNNs, IUPAC names encoded by language models bring in a rich amount of prior knowledge by basic words representing functional groups, without the need for sophisticated graph hyperparameter design.

\section{Method}
\begin{figure}[t]
\centering
\includegraphics[width=0.48\textwidth]{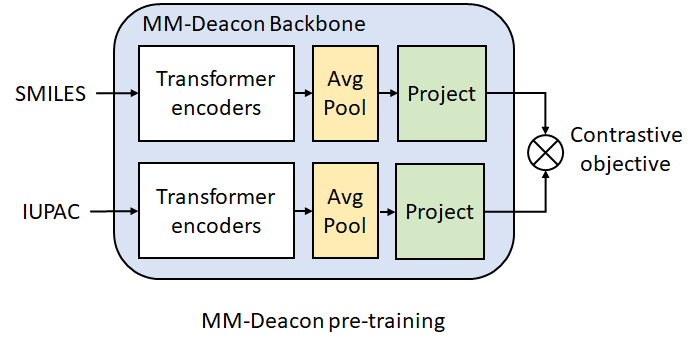}
\caption{Schematic diagram for MM-Deacon pre-training. SMILES and IUPAC are encoded by separate Transformers. Embeddings from encoders are average-pooled globally and projected to a joint chemical embedding space, where contrastive objectives are used to maximize mutual information for SMILES and IUPAC from the same molecule and distinguish SMILES and IUPAC from different molecules. \label{fig:backbone}}
\end{figure}
MM-Deacon is a deep neural network designed for SMILES-IUPAC joint learning with the goal of contrasting positive SMILES-IUPAC pairs from negative pairs and thus maximizing mutual information across different molecule languages. SMILES and IUPAC for the same molecule are regarded as positive pairs, while SMILES and IUPAC for different molecules are considered negative. Transformer encoders with multi-head self-attention layers are utilized to encode SMILES and IUPAC strings. Embeddings from the encoders are pooled globally and projected to the joint chemical embedding space. MM-Deacon is pre-trained on a dataset of 10 million molecules chosen at random from PubChem.
\subsection{Tokenizer}
We use a Byte-Pair Encoding (BPE) tokenizer for SMILES tokenization, as is shown by Chithrananda \textit{et al.} \citeyearpar{chithrananda2020chemberta} that BPE performed better than regex-based tokenization for SMILES on downstream tasks. For IUPAC name tokenization, a rule-based regex \cite{krasnov2021struct2iupac} that splits IUPAC strings based on suffixes, prefixes, trivial names, and so on is employed. The input sequence length statistics as well as the top 20 most frequent tokens in the SMILES and IUPAC corpora are displayed in Figs. \ref{fig:stat1} and \ref{fig:tokendis} (Appendix \ref{app:data}).
\subsection{Model architecture}
As illustrated in Fig. \ref{fig:backbone}, MM-Deacon takes SMILES and IUPAC strings as the input to separate branches. The input text string $s$ is tokenized and embedded into a numeric matrix representation $x$ within each branch, and the order of the token list is preserved by a positional embedding $p_x$. Then $x$ and $p_x$ are ingested by an encoder block $\varphi$ that consists of 6 layers of Transformer encoder. A Transformer encoder has two sub-layers, a multi-head attention layer and a fully-connected feed-forward layer. Each sub-layer is followed by a residual connection and layer normalization to normalize input values for all neurons in the same layer \cite{vaswani2017attention,ba2016layer}. The multi-head attention layer acquires long-dependency information by taking all positions into consideration. We then use a global average pooling layer $\rho$ to integrate features at all positions and a projection layer $\phi$ to project the integrated feature vector to the joint embedding space. Thus the final embedding $z$ of $x$ can be expressed as, 

\begin{equation}
z_{(x)}=\phi (\rho (\varphi(x+p_x))) \ .
\end{equation}

The maximum input token sequence length is set to 512. For each of the 6 Transformer encoder layers, we choose the number of self-attention heads as 12 and hidden size of 768. The projection layer $\phi$ projects the vector from length of 768 to 512 to make the representation more compact. Thus $z_{(x)}\in \mathbb{R}^{512}$.

\subsubsection{Contrastive loss}
Our goal is to align pairs of language modalities in the joint embedding space by maximizing mutual information of positive pairs and distinguishing them from negative pairs. For this purpose, we use InfoNCE \cite{oord2018representation,alayrac2020self,radford2021learning} as the contrastive loss. We do not construct negative pairs manually. Instead, during training, we obtain negative pairs in minibatches. Using a minibatch of $N$ SMILES-IUPAC pairs from $N$ molecules as input, $N$ positive pairs and $N^2-N$ negative pairs can be generated within the correlation matrix of $N$ SMILES strings and $N$ IUPAC strings. More specifically, the only positive pair for $i$-th SMILES is $i$-th IUPAC, while the remaining $N-1$ IUPAC strings form negative pairs with $i$-th SMILES. Therefore, the InfoNCE loss for $i$-th SMILES is,
\begin{equation}
L^{sl}_i=-\text{log }(\frac{\text{exp}(\text{sim}(z^{sl}_i,z^{ip}_i)/\tau)}{\sum_{j=1}^{N}\text{exp}(\text{sim}(z^{sl}_i,z^{ip}_j)/\tau)}) \ ,
\end{equation}

where $sl$ and $ip$ represent SMILES and IUPAC respectively. $\text{sim}()$ is the pairwise similarity function that employs cosine similarity in this work. $\tau$ is the temperature. Likewise, the loss function for $i$-th IUPAC is,

\begin{equation}
L^{ip}_i=-\text{log}(\frac{\text{exp}(\text{sim}(z^{sl}_i,z^{ip}_i)/\tau)}{\sum_{j=1}^{N}\text{exp}(\text{sim}(z^{sl}_j,z^{ip}_i)/\tau)}) \ .
\end{equation}

As a result, the final loss function is as follows,

\begin{equation}
L=\frac{1}{2N}\sum_{t\in \{sl,ip\}}\sum_{i=1}^{N}L^t_i \ .
\end{equation}

We pre-train MM-Deacon on 80 V100 GPUs for 10 epochs (15 hours in total) with a 16 batch size on each GPU using AdamW optimizer with a learning rate of $10^{-6}$. The temperature $\tau$ is set as 0.07 as in \cite{oord2018representation}.

\begin{figure}[t]
\centering
\includegraphics[width=0.48\textwidth]{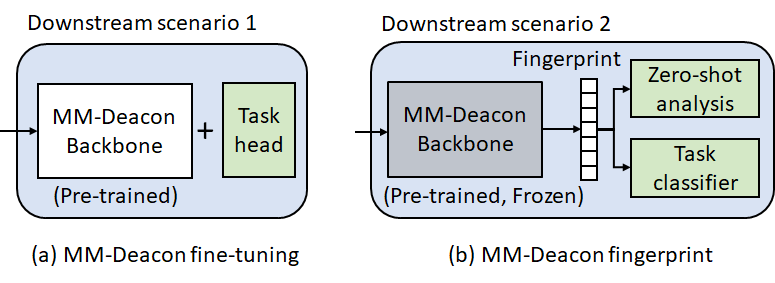}
\caption{Possible scenarios in the downstream stage. (a) MM-Deacon fine-tuning: A task-specific classification/regression head is attached to pre-trained MM-Deacon and optimized together on downstream task datasets. (b) MM-Deacon fingerprint: Pre-trained MM-Deacon is frozen. An input molecule is embedded as MM-Deacon fingerprint for zero-shot explorations (such as clustering analysis and similarity retrieval) and supervised tasks with the help of an extra classifier. \label{fig:downstream}}
\end{figure}

\begin{subsection}{Downstream stage}
Knowledge gained during pre-training can be transferred to downstream tasks in different ways. Fig. \ref{fig:downstream} lists two situations that make use of pre-trained MM-Deacon in the downstream stage.

\vspace{1mm}
\noindent \textbf{MM-Deacon fine-tuning:} A task-specific classification/regression head can be attached to pre-trained MM-Deacon and the system as a whole can be tuned on downstream task datasets.   

\vspace{1mm}
\noindent \textbf{MM-Deacon fingerprint:} Pre-trained MM-Deacon is frozen. An input molecule is embedded as MM-Deacon fingerprint for zero-shot explorations (such as clustering analysis and similarity retrieval) and supervised tasks with the help of an extra classifier.

\end{subsection}

\section{Experiments}
MM-Deacon was evaluated on seven molecular property prediction tasks from MoleculeNet benchmark \cite{wu2018moleculenet}, zero-shot cross-lingual retrieval, and a drug-drug interaction (DDI) prediction task.

\begin{subsection}{Molecular property prediction}
MoleculeNet benchmark provides a unified framework for evaluating and comparing molecular machine learning methods on a variety of molecular property prediction tasks ranging from molecular quantum mechanics to physiological themes, and is widely acknowledged as the standard in the research community \cite{hu2019strategies,chithrananda2020chemberta, xue2021x, zhu2021dual, wang2021molclr, liu2021pre}. Four classification datasets and three regression datasets from the MoleculeNet benchmark were utilized to evaluate our approach. 

\vspace{1mm}
\noindent \textbf{Data.} The blood-brain barrier penetration (BBBP), clinical trail toxicity (ClinTox), HIV replication inhibition (HIV), and side effect resource (SIDER) datasets are classification tasks in which molecule SMILES strings and their binary labels are provided in each task. Area Under Curve of the Receiver Operating Characteristic curve (ROC-AUC) is the performance metric in which the higher the value, the better the performance. For datasets with multiple tasks like SIDER, the averaged ROC-AUC across all tasks under the same dataset is reported. The fractions of train/val/test sets for each classification task are 0.8/0.1/0.1 with Scaffold split. Note that data split using molecule scaffolds (two-dimensional structural frameworks) results in more structurally distinct train/val/test sets, making it more challenging than random split \cite{wu2018moleculenet}. The water solubility data (ESOL), free solvation (FreeSolv), and experimental results of octabol/water distribution coefficient (Lipophilicity) datasets are all regression tasks to predict numeric labels given molecule SMILES strings. Root Mean Square Error (RMSE) is used as the evaluation metric in which the lower the value, the better the performance. As recommended by MoleculeNet, random split that divides each dataset into 0.8/0.1/0.1 for train/val/test sets is employed. The results on validation set are used to select the best model. To maintain consistency with MoleculeNet, we ran each task three times, each time with a different data split seed, to obtain the mean and standard deviation (std) of the metric. Details of each dataset such as the number of tasks and molecules it contains are displayed in Table \ref{tab:benchdata}. 

\begin{table}[h]
\centering
\fontsize{7.5}{11}\selectfont
\caption{MoleculeNet benchmark datasets used in this study. For each dataset, the number of molecules contained, the number of tasks it has, dataset split method, and evaluation metric type are listed. The first section in the table consists of classification tasks, and the second section consists of regression tasks. \label{tab:benchdata}}
\begin{tabular}{@{}cccccc@{}}
\toprule
Dataset & \# Molecules & \# Tasks & Split & Metric\\ \hline
BBBP  	& 2039 	& 1 & Scaffold & ROC-AUC	\\
ClinTox  	& 1478 	& 2 & Scaffold& ROC-AUC\\
HIV  	& 41127 	& 1 & Scaffold& ROC-AUC\\
SIDER  	& 1427 	& 27 & Scaffold& ROC-AUC\\ \hline
ESOL  	& 1128 	& 1 & Random & RMSE\\
FreeSolv  	& 642 	& 1 & Random & RMSE\\
Lipophilicity  	& 4200 	& 1 & Random & RMSE\\ \bottomrule
\end{tabular}
\end{table}

\vspace{1mm}
\noindent \textbf{Model.} We utilized the model shown in Fig. \ref{fig:downstream}(a) in which a linear layer serving as the task-specific head was added to pre-trained MM-Deacon SMILES branch for fine-tuning (IUPAC branch was removed). Cross-entropy loss was employed for classification tasks and MSE loss was employed for regression tasks. Hyperparameter tuning was performed using grid search with possible choices listed in Table \ref{hyper} (Appendix \ref{app:tune}). Each task was optimized individually.

\begin{table*}[!htb]
\centering
\fontsize{7.5}{11}\selectfont
\begin{tabular}{cccccccc}
\toprule
\textbf{Method}& \textbf{BBBP} & \textbf{ClinTox} & \textbf{HIV}  &\textbf{SIDER} &\textbf{ESOL} &\textbf{FreeSolv}&\textbf{Lipophilicity}\\ \hline
RF & 71.4$\pm$0.0  & 71.3$\pm$5.6& 78.1$\pm$0.6 & 68.4$\pm$0.9&1.07$\pm$0.19 &2.03$\pm$0.22 &0.876$\pm$0.040 \\ 
KernelSVM&72.9$\pm$0.0&66.9$\pm$9.2&79.2$\pm$0.0&68.2$\pm$1.3&-&-&-\\
Multitask& 68.8$\pm$0.5& 77.8$\pm$5.5& 69.8$\pm$3.7& 66.6$\pm$2.6& 1.12$\pm$0.15& 1.87$\pm$0.07& 0.859$\pm$0.013 \\
GC& 69.0$\pm$0.9& 80.7$\pm$4.7& 76.3$\pm$1.6& 63.8$\pm$1.2& 0.97$\pm$0.01& 1.40$\pm$0.16 & 0.655$\pm$0.036\\
Weave& 67.1$\pm$1.4& 83.2$\pm$3.7& 70.3$\pm$3.9& 58.1$\pm$2.7& 0.61$\pm$0.07 & 1.22$\pm$0.28 & 0.715$\pm$0.035\\
MPNN& -&-&-&-& 0.58$\pm$0.03& 1.15$\pm$0.12& 0.719$\pm$0.031 \\ \hline
Hu \textit{et al.} \citeyearpar{hu2019strategies} & 70.8$\pm$1.5    & 78.9$\pm$2.4     & 80.2$\pm$0.9 & 65.2$\pm$0.9&-&-&-   \\
MolCLR \cite{wang2021molclr}& 73.6$\pm$0.5&93.2$\pm$1.7 &80.6$\pm$1.1 & 68.0$\pm$1.1&-&-&-   \\ 
DMP \cite{zhu2021dual}& 78.1$\pm$0.5&95.0$\pm$0.5  & \textbf{81.0$\pm$0.7}&69.2$\pm$0.7&-&-&- \\ 
X-Mol \cite{xue2021x} & \textbf{96.2$\pm$N/A}  &98.4$\pm$N/A& 79.8$\pm$N/A &-&0.578$\pm$N/A & 1.108$\pm$N/A& \textbf{0.596$\pm$N/A}\\
GraphMVP \cite{liu2021pre} & 72.4$\pm$1.6  &77.5$\pm$4.2& 77.0$\pm$1.2 &63.9$\pm$1.2 &1.029$\pm$N/A&-& 0.681$\pm$N/A\\
MLM-\texttt{[CLS]}& 70.6$\pm$4.5&93.2$\pm$0.1  & 77.9$\pm$0.2&64.8$\pm$1.3&0.640$\pm$0.023&1.21$\pm$0.046&0.804$\pm$0.037 \\ 
MM-Deacon & 78.5$\pm$0.4&\textbf{99.5$\pm$0.3}& 80.1$\pm$0.5 &\textbf{69.3$\pm$0.5} &\textbf{0.565$\pm$0.014}&\textbf{0.926$\pm$0.013}&0.650$\pm$0.021\\ \bottomrule
\end{tabular}
\caption{\label{tab:benchresult}
Results in terms of mean and std for each dataset included from MoleculeNet benchmark. The first section of the table is the results imported from the MoleculeNet paper \cite{wu2018moleculenet}. The second section lists results from cutting-edge molecular pre-training and fine-tuning approaches together with MM-Deacon. MLM-\texttt{CLS} is the model that uses the same Transformer settings as MM-Deacon SMILES branch, pre-trained with MLM loss on 10M molecules, and fine-tuned through \texttt{[CLS]} token with the same downstream setting as MM-Deacon. The best outcome is denoted by \textbf{bold}.}
\end{table*}

\vspace{1mm}
\noindent \textbf{Results.} Table \ref{tab:benchresult} shows the mean and std results for each dataset. The first half of the table displays results imported from MoleculeNet \cite{wu2018moleculenet}, while the second section shows the results from MM-Deacon and other state-of-the-art molecular pre-training and fine-tuning approaches. MLM-\texttt{[CLS]} denotes our implementation of a Chemberta \cite{chithrananda2020chemberta} variant that uses the same Transformer settings as MM-Deacon SMILES branch, pre-trained with MLM loss on 10M molecules, and fine-tuned through \texttt{[CLS]} token with the same downstream setting as MM-Deacon. MM-Deacon exceeds the performance of traditional machine learning methods like random forest (RF) and task-specific GNNs reported in MoleculeNet work by a significant margin for most of the tasks. When compared to other pre-training based approaches, MM-Deacon outperforms the existing state-of-the-art approaches in four of the seven datasets and is comparable in the remaining three, with major improvements on ClinTox and FreeSolv.

All pre-training based methods were pre-trained on millions of molecules, with the exception of GraphMVP, which was pre-trained on 50K molecules. The requirement that molecules have both 2D and 3D structure information available at the same time to be qualified has limited the scalability of GraphMVP. MM-Deacon and MLM-\texttt{CLS} both used 6 layers of Transformer blocks to process SMILES. For each task, MM-Deacon, which was pre-trained with both SMILES and IUPAC, outscored MLM-\texttt{CLS}, which was pre-trained with SMILES only. MM-Deacon and DMP performed comparably on the four classification tasks, while DMP used 12 layers of Transformer blocks for SMILES and a 12-layer GNN to encode a molecule 2D graph, which is nearly twice the size of MM-Deacon model.

Moreover, we found that BBBP test set is significantly more challenging than the validation set, which is consistent with the results published in the MoleculeNet paper \cite{wu2018moleculenet}. The substantially high accuracy X-Mol achieved on the BBBP dataset could be due to either the 1.1 billion molecules they utilized for pre-training or a different dataset division approach they employed.
\end{subsection}

\begin{subsection}{Zero-shot cross-lingual retrieval}
In addition to conducting fine-tuning on supervised tasks like molecular property prediction, pre-trained MM-Deacon can be employed directly in large-scale zero-shot analysis. Zero-shot cross-lingual retrieval operates on top of MM-Deacon fingerprint generated by pre-trained MM-Deacon given molecule SMILES or IUPAC as input. This task enables the retrieval of similar molecules across languages without the need for translation, and it can also be used to evaluate the learned agreement in the joint embedding space between SMILES and IUPAC representations. 

\vspace{1mm}
\noindent \textbf{Data.} 100K molecules were randomly chosen from PubChem dataset after excluding the 10 million molecules used for MM-Deacon pre-training. SMILES and IUPAC strings are provided for each molecule. We used average recall at K (R@1 and R@5) to measure the percentage of the ground truth that appears in the top K retrieved molecules.

\vspace{1mm}
\noindent \textbf{Model.} Pre-trained MM-Deacon was used for MM-Deacon fingerprint generation, as shown in Fig. \ref{fig:downstream}(b). As a result, each SMILES and IUPAC string was encoded as MM-Deacon SMILES fingerprint and IUPAC fingerprint respectively. Cosine similarity between a query and molecules in the search candidates was used to determine the ranking.

\vspace{1mm}
\noindent \textbf{Results.} Fig. \ref{fig:search} shows the outcomes of SMILES-to-IUPAC and IUPAC-to-SMILES retrieval in terms of recall. We not only performed retrieval directly on the entire 100K molecules, but also reported the results on smaller groups of molecules (100, 10K) to get a more thorough picture of the retrieval performance. MM-Deacon gets a R@5 above 85\% for both types of cross-lingual retrieval even while executing retrieval on 100K molecules. Moreover, Figs. \ref{fig:s2i} and \ref{fig:i2s} show an example of SMILES-to-IUPAC retrieval and an example of IUPAC-to-SMILES retrieval respectively.

\begin{figure}[!tb]
\centering
\includegraphics[width=0.48\textwidth]{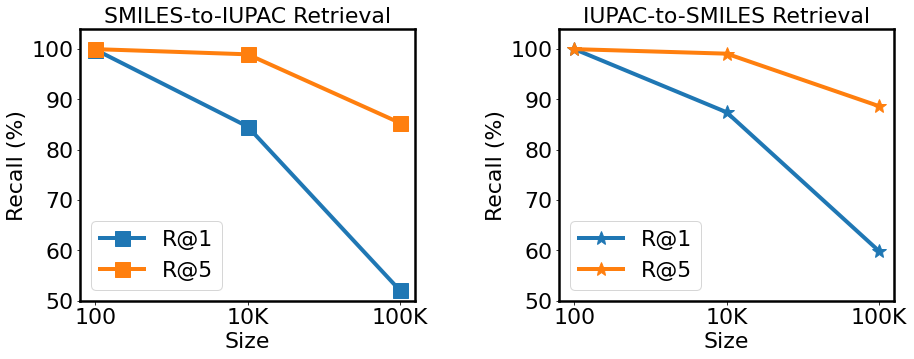}
\caption{Average recall for cross-lingual retrieval on groups of molecules with different sizes. \label{fig:search}}
\end{figure}

\begin{figure}[!tb]
\centering
\includegraphics[width=0.48\textwidth]{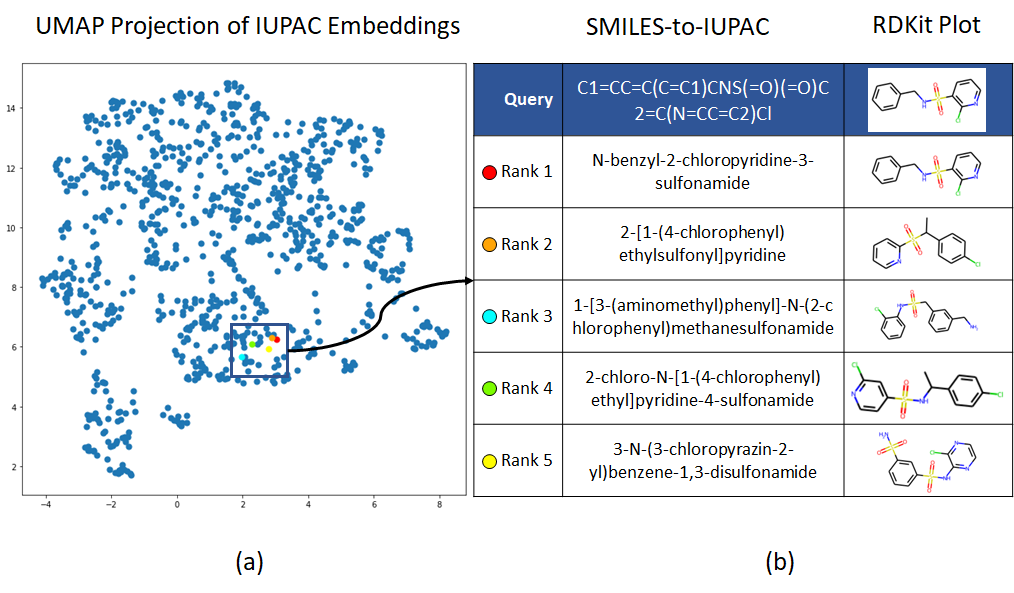}
\caption{An example of SMILES-to-IUPAC retrieval on 100K molecules. (\textbf{a}) 2D projection of IUPAC fingerprints using UMAP \cite{mcinnes2018umap} for top 1K ranked molecules. (\textbf{b}) SMILES query and top 5 IUPAC names, with RDKit plots aside for interpretation purpose. The black square in (\textbf{a}) marks the top 5 IUPAC names in (\textbf{b}). \label{fig:s2i}}
\end{figure}

\begin{figure}[!tb]
\centering
\includegraphics[width=0.48\textwidth]{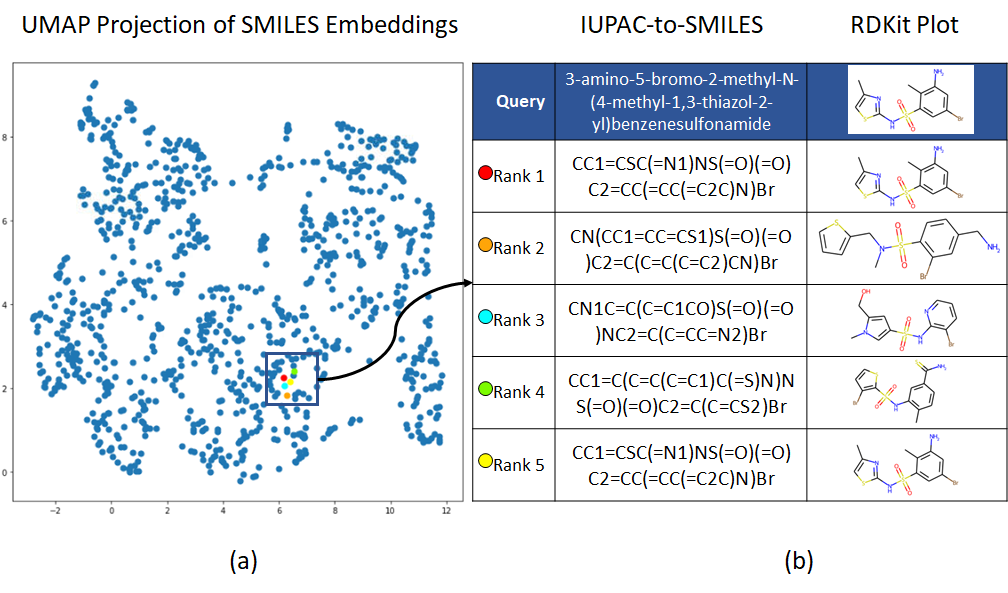}
\caption{An example of IUPAC-to-SMILES retrieval on 100K molecules. (\textbf{a}) 2D projection of top 1K ranked SMILES fingerprints using UMAP. (\textbf{b}) IUPAC query and top 5 SMILES strings, with RDKit plots aside for interpretation purpose. The black square in (\textbf{a}) marks the top 5 SMILES strings in (\textbf{b}). \label{fig:i2s}}
\end{figure}

Additional retrieval examples for scenarios where the performance is difficult to be quantified, such as retrieval queried by a free combination of tokens and unilingual retrieval are included in Appendix \ref{app:retrieval}. 

\end{subsection}

\begin{subsection}{DDI prediction}
\begin{table}
\centering
\fontsize{7.5}{11}\selectfont
\caption{DDI prediction results using 5-fold cross-validation. * denotes results cited from \cite{zhang2017predicting}. $^\ddagger $  marks results using drug structural features. Ensemble methods used drug features obtained from substructure, target, enzyme, transporter, pathway, indication, and off side effect properties. Methods using embeddings derived from pre-trained models and an MLP classifier for classification are denoted by \textsuperscript{\textdagger}. \label{tab2}}.
\begin{tabular}{@{}ccccc@{}}
\toprule
Method  & AUC   & AUPR  & Precision & Recall \\ \hline
Neighbor recommender*$^\ddagger $                                       & 0.936 & 0.759 & 0.617     & 0.765      \\ 
Random walk*$^\ddagger $& 0.936 & 0.758 & 0.763     & 0.616     \\ 
Ensemble method (L1)*  & \textbf{0.957} & 0.807 & 0.785     & 0.670      \\ 
Ensemble method (L2)*    & 0.956 & 0.806 & 0.783     & 0.665    \\ 
DPDDI \cite{feng2020dpddi} & 0.956 & 0.907 & 0.754     & 0.810  \\ 
MLM-\texttt{[CLS]}\textsuperscript{\textdagger} & 0.943 & 0.901 & 0.784     & 0.813 \\ \hline
MM-Deacon (SMILES)\textsuperscript{\textdagger}                                                            & 0.946 & 0.911 & 0.805     & 0.823      \\ 
MM-Deacon (IUPAC)\textsuperscript{\textdagger}& 0.947 & 0.913 & \textbf{0.834}     & 0.797 \\ 
MM-Deacon (concat)\textsuperscript{\textdagger} & 0.950 & \textbf{0.918} & 0.819     & \textbf{0.824}\\ \bottomrule
\end{tabular}
\end{table}
The effectiveness of combining MM-Deacon fingerprints with a task-specific classifier for supervised learning was tested on a DDI prediction task. The objective of this task is to predict whether or not any two given drugs have an interaction.

\vspace{1mm}
\noindent \textbf{Data.} The DDI dataset \cite{zhang2017predicting} used here includes 548 drugs, with 48,584 known interactions, and 101,294 non-interactions (may contain undiscovered interactions at the time the dataset was created). We obtained the SMILES and IUPAC names for each drug from PubChem. Stratified 5-fold cross-validation with drug combination split was utilized. The evaluation metrics are Area Under the ROC Curve (AUC), Area Under the Precision-Recall Curve (AUPR), precision, and recall, with AUPR serving as the primary metric \cite{zhang2017predicting}.

\vspace{1mm}
\noindent \textbf{Model.} MM-Deacon fingerprints of paired drugs are concatenated and fed into a multi-layer perceptron (MLP) network implemented by scikit-learn \cite{scikit-learn} for binary classification. Three different types of fingerprints are used for MM-Deacon: SMILES, IUPAC, and concatenated SMILES and IUPAC fingerprints. The MLP has one hidden layer with 200 neurons. ReLU activation and a learning rate of $10^{-3}$ are used. 

\vspace{1mm}
\noindent \textbf{Results.} As shown in Table \ref{tab2}, MM-Deacon outperforms other methods in terms of AUPR, precision and recall, with the maximum AUPR obtained when SMILES and IUPAC fingerprints were concatenated as input feature set. Ensemble models \cite{zhang2017predicting} included extra bioactivity related features in addition to drug structural properties. DPDDI \cite{feng2020dpddi} encoded molecule graph with GNNs, from which latent features were concatenated for pairs of drugs and ingested into a deep neural network.

Table \ref{tab:drugdis} shows the top 20 most potential interactions predicted by MM-Deacon (concat) in the non-interaction set (false positives), 13 out of which are confirmed as true positives by DrugBank\footnote{https://go.drugbank.com/drug-interaction-checker}. While, the number is 7/20 for ensemble models \cite{zhang2017predicting}. 

\begin{table}
\centering
\fontsize{7.5}{11}\selectfont
\setlength{\tabcolsep}{2pt}
\caption{Top 20 most potential interactions in the non-interaction set. True interactions verified by DrugBank are marked in \textbf{bold}.\label{tab:drugdis}}
\begin{tabular}{@{}cll@{}}
\toprule
Rank    & Drug 1 ID (Name)   & Drug 2 ID (Name)\\ \hline
1&\textbf{DB00722 (Lisinopril)} &\textbf{DB00915 (Amantadine)} \\
2&\textbf{DB00213 (Pantoprazole)} &\textbf{DB00310 (Chlorthalidone)} \\
3&\textbf{DB00776 {\tiny (Oxcarbazepine)}}& \textbf{DB00433 (Prochlorperazine)} \\
4&DB00481 (Raloxifene)& DB00501 (Cimetidine) \\
5&\textbf{DB01193 (Acebutolol)} &\textbf{DB00264 (Metoprolol)} \\
6&DB00250 (Dapsone)& DB00230 (Pregabalin) \\
7&\textbf{DB00415 (Ampicillin)} &\textbf{DB01112 (Cefuroxime)} \\
8&\textbf{DB00582 (Voriconazole)} &\textbf{DB01136 (Carvedilol)} \\
9&\textbf{DB01079 (Tegaserod)} &\textbf{DB00795 (Sulfasalazine)} \\
10&\textbf{DB01233 {\tiny (Metoclopramide)}} &\textbf{DB00820 (Tadalafil)} \\
11&DB00213 (Pantoprazole) &DB00513 (Aminocaproic Acid) \\
12&DB01195 (Flecainide) &DB00584 (Enalapril)\\
13&\textbf{DB00758 (Clopidogrel)} &\textbf{DB01589 (Quazepam)} \\
14&\textbf{DB01136 (Carvedilol)} &\textbf{DB00989 (Rivastigmine)} \\
15&\textbf{DB00586 (Diclofenac)} &\textbf{DB01149 (Nefazodone)} \\
16&DB00407 (Ardeparin) &DB00538 (Gadoversetamide) \\
17&\textbf{DB00203 (Sildenafil)} &\textbf{DB00603 {\tiny (Medroxyprogesterone Acetate)}} \\
18&DB00601 (Linezolid) &DB01112 (Cefuroxime) \\
19&\textbf{DB00275 (Olmesartan)} &\textbf{DB00806 (Pentoxifylline)} \\
20&DB00231 (Temazepam) & DB00930 (Colesevelam)\\ \bottomrule
\end{tabular}
\end{table}
\end{subsection}

\section{Discussions}
After being pre-trained on 10 million molecules, MM-Deacon showed outstanding knowledge transfer capabilities to various downstream scenarios (Fig. \ref{fig:downstream}) where a pre-trained model could be used. The competitive performance on seven molecular property prediction tasks from MoleculeNet benchmark demonstrated the effectiveness of the pre-trained MM-Deacon when adopting a network fine-tuning scheme as shown in Fig. \ref{fig:downstream}(a). The evaluation results of zero-shot cross-lingual retrieval further revealed that MM-Deacon SMILES and IUPAC fingerprints shared a substantial amount of mutual information, implying that an IUPAC name can be used directly without first being translated to SMILES format as chemists have done in the past. The DDI prediction task showed that MM-Deacon also allows directly using embeddings in the joint cross-modal space as molecular fingerprints for downstream prediction tasks, which is a widely used strategy in cheminformatics. 

MM-Deacon profited from the alignment of two molecule languages with distinct forms of nomenclatures, as opposed to the baseline MLM-\texttt{[CLS]} model, which was pre-trained on SMILES representation only. Furthermore, we looked at molecule-level and token-level alignments of MM-Deacon to untangle the outcome of cross-lingual contrastive learning.

\begin{figure}[!tb]
\centering
\includegraphics[width=0.48\textwidth]{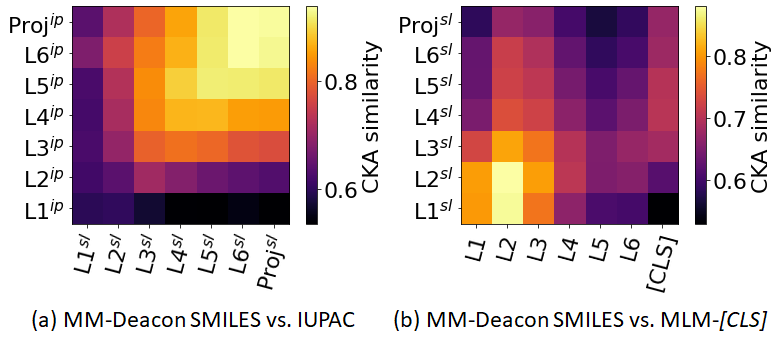}
\caption{Representation comparison using CKA. $sl$ and $ip$ denote SMILES and IUPAC respectively. $L$ represents a Transformer layer. (a) Comparison between MM-Deacon SMILES and IUPAC branches. (b) Comparison between MM-Deacon SMILES branch and MLM-\texttt{[CLS]}. \label{fig:compare}}
\end{figure}

\begin{subsection}{Molecule-level alignment}
We used centered kernel alignment (CKA) \cite{kornblith2019similarity} with RBF kernel to compare representations between different layers. In Fig. \ref{fig:compare}(a), the representations of 6 Transformer layers and the final projection layer were compared between MM-Deacon SMILES and IUPAC branches, where the representations differ in shallow layers, while reach a high level of alignment in deeper layers. In Fig. \ref{fig:compare}(b), both the MM-Deacon SMILES branch and MLM-\texttt{[CLS]} model take SMILES as the input, therefore the shallow layers have a high alignment score, while the representation varies as the network grows deeper. Fig. \ref{fig:compare} shows that MM-Deacon aligned SMILES and IUPAC representations effectively, and that molecular representations trained with SMILES and IUPAC differs from representations trained only on SMILES.
\end{subsection}

\subsection{Token-level alignment}
\begin{figure}[!tb]
\centering
\includegraphics[width=0.48\textwidth]{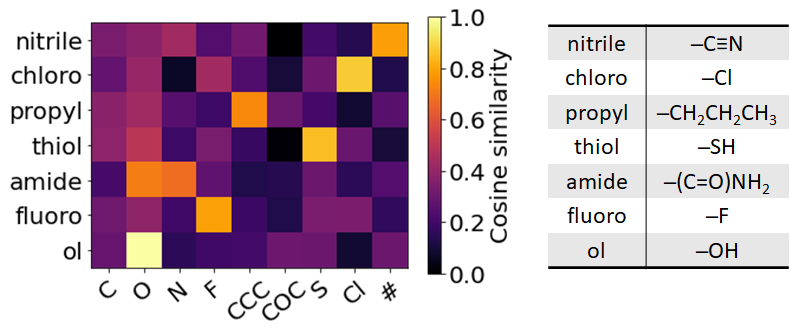}
\caption{Token-level alignment between SMILES and IUPAC. Left: heat map of MM-Deacon fingerprint cosine similarity between SMILES and IUPAC tokens. Right: IUPAC token correspondence in SMILES representation.  \# represents a triple bond in SMILES language. \label{fig:token}}
\end{figure}

The cosine similarity matrix of MM-Deacon fingerprints between tokens from the IUPAC corpus and tokens from the SMILES corpus is shown in Fig. \ref{fig:token}. The table in Fig. \ref{fig:token} lists IUPAC tokens expressed in SMILES language, and the heat map demonstrates that there exists a good token-level alignment between SMILES and IUPAC.

\section{Conclusion}
In this study, we proposed a novel method for multilingual molecular representation learning that combines mutual information from SMILES-IUPAC joint training with a self-supervised contrastive loss. We evaluated our approach for molecular property prediction, zero-shot cross-lingual retrieval, and DDI prediction. Our results demonstrate that the self-supervised multilingual contrastive learning framework holds enormous possibilities for chemical domain exploration and drug discovery. In future work, we plan to scale MM-Deacon pre-training to larger dataset sizes, as well as investigate the applicability of MM-Deacon to other types of molecule languages.
\section*{Acknowledgements}
We would like to thank Min Xiao and Brandon Smock for some insightful discussions.


\bibliography{custom}
\bibliographystyle{acl_natbib}

\newpage

\appendix
\section{Data statistics}
\label{app:data}
\begin{figure}[h]
\centering
\includegraphics[width=0.48\textwidth]{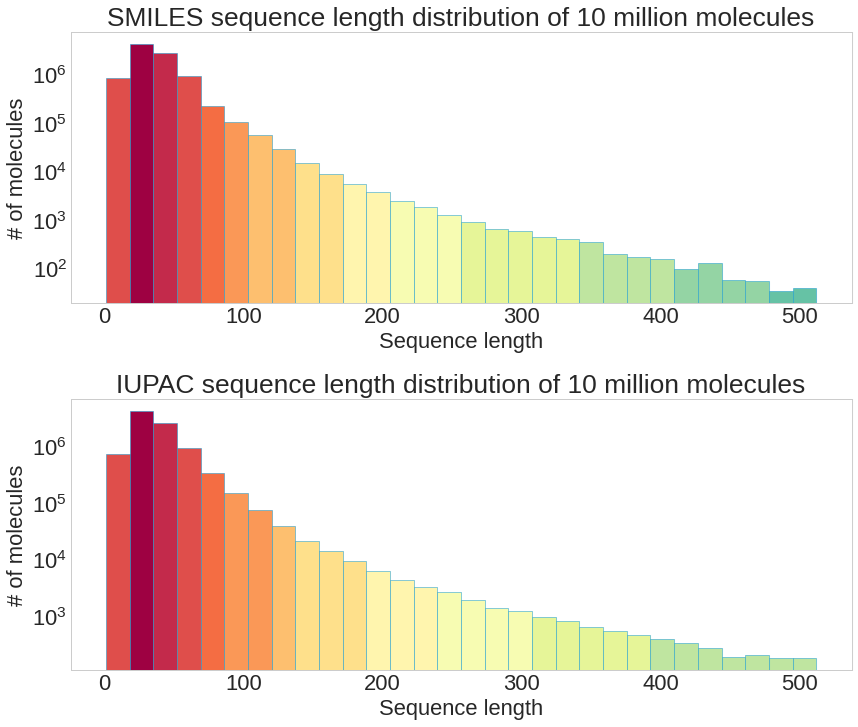}
\caption{Distributions of SMILES and IUPAC sequence lengths in the 10-million-molecule training set. Y axis shows the number of molecules in log scale. \label{fig:stat1}}
\end{figure}

The distributions of SMILES and IUPAC sequence lengths in the training set are shown in Fig. \ref{fig:stat1} in log scale for the y axis. 

\begin{figure}[h]
\centering
\includegraphics[width=0.48\textwidth]{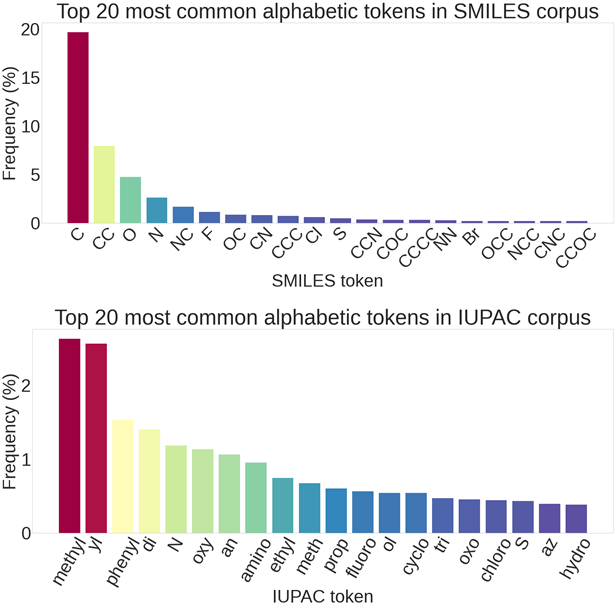}
\caption{Schematic illustration of differences in SMILES and IUPAC. \label{fig:tokendis}}
\end{figure}

Fig. \ref{fig:tokendis} displays the top 20 most frequent alphabetic tokens in SMILES and IUPAC corpora. Token \textit{C}, which simply denotes a carbon atom, appears nearly 20\% of the time in SMILES language. On the other hand, the frequency of IUPAC tokens is quite evenly distributed, with the prefixes \textit{methyl} and \textit{phenyl} as well as the suffix \textit{yl} from the \textit{alkyl} functional group being the top 3 most common tokens. 

\section{Hyperparameter tuning}
\label{app:tune}
Table \ref{hyper} lists the search space of hyperparameter tuning for seven molecular property prediction tasks from MoleculeNet benchmark for MM-Deacon and MLM-\texttt{[CLS]}. We employed grid search to find the best hyperparameters. Each task was optimized individually.
\begin{table}[h]
\centering
\fontsize{10}{11}\selectfont
\caption{Grid search choices for molecular property prediction hyperparameter tuning for models MM-Deacon and MLM-\texttt{[CLS]}. \label{hyper}}.
\begin{tabular}{@{}cc@{}}
\toprule
Parameter  & Choices\\ \hline
Learning rate  & \{1e-6, 5e-6, 1e-5, 5e-5, 1e-4\} \\ 
Batch size  & \{2, 4, 8, 12, 24\} \\
Epochs & 20 \\ 
Early termination & Bandit policy \\ \bottomrule
\end{tabular}
\end{table}

\section{Extra zero-shot retrieval examples}
\label{app:retrieval}

\begin{figure}[h]
\centering
\includegraphics[width=0.48\textwidth]{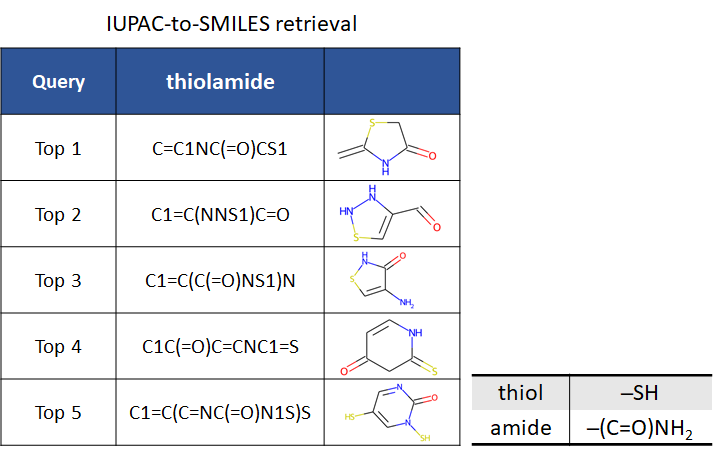}
\caption{An example of IUPAC-to-SMILES retrieval on 100K molecules with a free-form IUPAC query. Left: a table containing the IUPAC query and the top 5 retrieval molecules in SMILES representation, as well as RDKit plots for interpretation. Right: corresponding structure characteristics in SMILES language of IUPAC tokens \textit{thiol} and \textit{amide}. \label{fig:token_i2s}}
\end{figure}

An example of cross-lingual retrieval using a free-form IUPAC query is shown in Fig. \ref{fig:token_i2s}. The IUPAC query \textit{thiolamide}, a combination of tokens \textit{thiol} and \textit{amide}, does not exist in the IUPAC corpus (is not a substring of any IUPAC name). When searching on top of MM-Deacon fingerprints, all of the retrieved molecules have the features of atom \textit{S}, \textit{N} and \textit{C=O}. That is, the semantic meaning of the query is captured. 

\begin{figure}[t]
\centering
\includegraphics[width=0.48\textwidth]{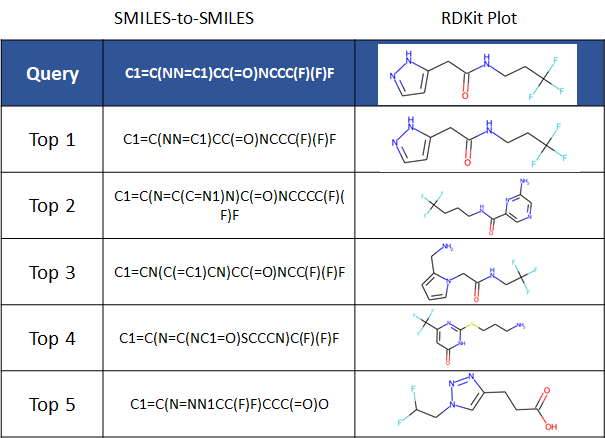}
\caption{An example of SMILES-to-SMILES retrieval on 100K molecules. The SMILES query, the top 5 retrieved SMILES names, and the corresponding RDKit plots are presented. \label{fig:uni_s2s}}
\end{figure}

\begin{figure}[t]
\centering
\includegraphics[width=0.48\textwidth]{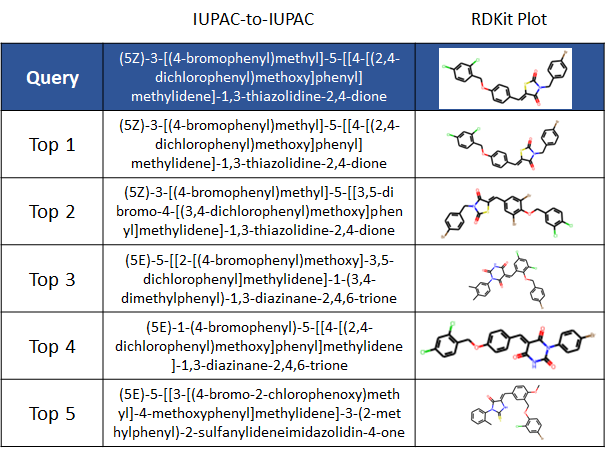}
\caption{An example of IUPAC-to-IUPAC retrieval on 100K molecules. The IUPAC query, the top 5 retrieved IUPAC names, and the corresponding RDKit plots are presented. \label{fig:uni_i2i}}
\end{figure}

In addition to cross-lingual retrieval, unilingual similarity retrieval is also supported, while its performance is difficult to be quantified. Figs. \ref{fig:uni_s2s} and \ref{fig:uni_i2i} show an example of SMILES-to-SMILES retrieval and IUPAC-to-IUPAC retrieval respectively using MM-Deacon fingerprints. 

\end{document}